\documentclass[10pt,twocolumn,letterpaper]{article}

\usepackage{wacv}              % To produce the CAMERA-READY version
\usepackage[T1]{fontenc}
\usepackage[utf8]{inputenc}

\definecolor{wacvblue}{rgb}{0.21,0.49,0.74}
\usepackage[pagebackref,breaklinks,colorlinks,allcolors=wacvblue]{hyperref}
\usepackage{multirow}
\def\wacvPaperID{ 1536} % *** Enter the WACV Paper ID here
\def\confName{WACV}
\def\confYear{2027}

\title{ORCA: Occlusion-Aware Refinement and Completion for Novel View Synthesis}

\author{
Weronika Jakubowska$^{1}$ \quad
Maciej Zięba$^{1}$ \quad
Przemysław Spurek$^{2,3}$\\
$^{1}$Wrocław University of Science and Technology \quad
$^{2}$Jagiellonian University \quad
$^{3}$IDEAS Research Institute\\
{\tt\small weronika.jakubowska@pwr.edu.pl}
}

\begin{document}
\maketitle

\begin{abstract}
Novel-view synthesis from a single image is a fundamentally ambiguous problem. As the camera moves away from the input viewpoint, previously hidden regions become visible, exposing missing geometry and holes in the reconstructed scene. Existing methods often rely on generative models to complete such regions. However, many of these artifacts are small gaps near depth boundaries and do not require generating new scene content.

In order to eliminate expensive process of generating image we introduce ORCA, an occlusion-aware method for reconstructing and completing explorable 3D scenes from a single image. ORCA first introduces 3D structure into a Gaussian-anchor representation using monocular depth while preserving the original camera-ray correspondence. During scene exploration, missing regions are handled based on their size and structure. Small disocclusions are repaired using RGB-D information already available in the reconstruction, while generative inpainting is reserved for larger regions that cannot be reliably recovered from the scene. New Gaussian anchors are added and optimized locally without modifying the existing representation. By reducing unnecessary reliance on generative inpainting, ORCA limits generation-induced hallucinations and better preserves the content and structure of the original scene.

On DIV2K, ORCA improves novel-view quality over VistaDream across all reported metrics, increasing MUSIQ from 61.60 to 68.71 and CLIP-IQA from 0.474 to 0.574. These results show that many novel-view artifacts can be repaired effectively by reusing information already present in the reconstructed scene.

\begin{figure}[h]
    \centering
    \includegraphics[width=0.8\columnwidth]{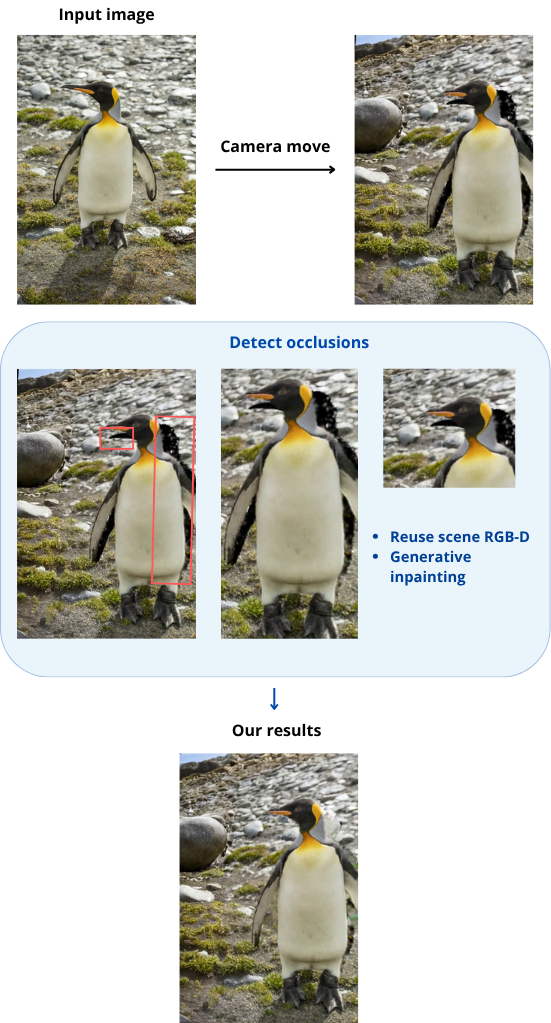}
    \caption{
        Camera motion reveals disoccluded regions that are not covered by the
        initial reconstruction. ORCA reuses existing scene information whenever
        possible and applies generative completion only to regions that cannot
        be reliably recovered from the reconstructed scene.
    }
    \label{fig:teaser}
\end{figure}

\end{abstract}   
\section{Introduction}

Reconstructing a 3D scene from a single image is difficult because the input provides only one view of the scene. Depth is ambiguous, and large parts of the scene may be hidden behind foreground objects. These limitations become visible as soon as the camera starts to move. A novel viewpoint can reveal incorrect geometry, gaps between surfaces, or regions that were never observed in the input image. Figure~\ref{fig:teaser} illustrates how camera motion exposes disoccluded regions and how ORCA repairs them using either existing scene information or generative completion.

Recent methods often address this problem by generating additional views or completing missing regions with generative models~\cite{vistadream,yu2025wonderworldinteractive3dscene,wang2026one2scenegeometricconsistentexplorable}. This makes it possible to extend the scene beyond the original observation, but generated content is not guaranteed to agree with the geometry that has already been reconstructed. A generated image may look plausible from one viewpoint while introducing inconsistencies when the scene is viewed from another.

We found that generation is not necessary for every region exposed by camera motion. Many artifacts that appear in novel views are small gaps around object boundaries. In such cases, the missing appearance is often already present in the reconstruction. The problem is instead that the existing geometry does not provide enough coverage from the new viewpoint. Larger occluded regions are different: they may contain content that was never visible and therefore cannot be recovered from the reconstructed scene alone.

This distinction motivates our approach, named ORCA, an Occlusion-Aware Refinement and Completion method for reconstructing explorable 3D scenes from a single image. ORCA uses information already present in the reconstruction whenever possible and introduces generated content only when it is needed. Small disocclusions are repaired using nearby RGB-D information from the reconstructed scene. We select background samples according to depth so that the added geometry is placed behind foreground objects rather than extending their surfaces into the missing region. Larger regions that cannot be recovered in this way are completed using generative inpainting.

The initial scene is represented using neural Gaussian anchors and is reconstructed from the input image together with an outpainted extension. Since both images correspond to the same camera viewpoint, they provide additional appearance information but no multi-view depth constraints. We therefore use monocular depth to introduce 3D structure. Gaussian positions are moved along their original camera rays according to the estimated relative depth, preserving their correspondence with the input view. The deformed geometry is then kept fixed while the remaining representation is fine-tuned.

Once the initial scene has been reconstructed, we render it along a camera trajectory and identify regions that become uncovered. Missing regions are repaired one at a time. New Gaussian anchors are added only inside the selected region and optimized without changing the existing scene representation. After each repair, the trajectory is rendered again before the next region is selected. This allows later repairs to take into account geometry that has already been added.

Unlike approaches that rely on generation whenever a new viewpoint reveals a hole, ORCA first checks whether the missing region can be recovered from the current reconstruction. This makes it possible to repair many disocclusions without synthesizing new appearance and limits generative completion to regions where the scene does not provide enough information.

Our main contributions are:
\begin{itemize}
\item We present a method for reconstructing a coherent and explorable 3D scene from a single image. Monocular depth is used to introduce 3D structure into a neural Gaussian-anchor representation while preserving the appearance observed in the input view.

\item We introduce an occlusion-repair mechanism that treats small geometric gaps differently from regions containing genuinely unseen content. Small disocclusions are reconstructed using RGB-D information already present in the scene, while generative completion is used only when the missing region cannot be recovered reliably from the existing representation.

\end{itemize}

\section{Related Work}

\subsection{Neural Scene Representations}

Neural Radiance Fields (NeRF) model a scene as a continuous function of density and view-dependent appearance and render novel views through volumetric integration~\cite{mildenhall2020nerfrepresentingscenesneural}. In contrast, 3D Gaussian Splatting (3DGS) represents the scene explicitly with anisotropic Gaussian primitives, enabling fast rendering while maintaining high visual quality~\cite{kerbl3Dgaussians}.

More recent approaches combine explicit geometric structures with learned neural features. Affine-Equivariant Kernel Space Encoding uses Gaussian kernels as a spatial support for locally transformable neural features~\cite{zielinski2026affineequivariantkernelspaceencoding}, while GaINeR combines trainable Gaussian primitives with a neural implicit image representation and demonstrates geometry-aware lifting and depth-guided editing~\cite{jakubowska2026gainergeometryawareimplicitnetwork}. Our method builds on IRIS~\cite{wilczynski2026irisintersectionawareraybasedimplicit}, which represents the scene using Gaussian neural anchors and aggregates their features through ray-Gaussian interactions. We use this representation as the basis for the depth deformation and disocclusion repair stages described in Sec.~\ref{sec:method}.

\subsection{Single-Image 3D Scene Reconstruction}

Reconstructing a 3D scene from a single image is inherently ambiguous because the input provides neither multi-view geometry nor observations of occluded regions. One common strategy is therefore to generate additional views and use them as supervision for reconstruction. SyncDreamer~\cite{liu2024syncdreamergeneratingmultiviewconsistentimages} generates multi-view consistent observations in an object-centric setting, while ZeroNVS~\cite{zeronvs} targets zero-shot novel-view synthesis of real scenes under larger viewpoint changes. CAT3D~\cite{gao2024cat3d} follows a generate-then-reconstruct strategy, where a set of generated views is first produced and then used to recover a 3D representation.

Several recent methods make the generation process more explicitly aware of camera geometry. MVGenMaster~\cite{cao2025mvgenmasterscalingmultiviewgeneration} incorporates camera parameters and 3D priors into multi-view diffusion, Stable Virtual Camera~\cite{zhou2025stablevirtualcameragenerative} generates views conditioned on target camera trajectories, and 3D-Adapter~\cite{chen20253dadaptergeometryconsistentmultiviewdiffusion} introduces geometric feedback during multi-view generation. Director3D~\cite{li2024director3drealworldcameratrajectory} explores a related setting in which scene content and camera trajectories are generated jointly.

A second line of work focuses more directly on constructing explorable 3D scenes. WonderWorld~\cite{yu2025wonderworldinteractive3dscene} combines image outpainting with geometric initialization and depth guidance to progressively extend the reconstructed scene. ExScene~\cite{gong2025exscenefreeview3dscene} first constructs a global scene representation from generated appearance and depth and then refines it using generative priors. One2Scene~\cite{wang2026one2scenegeometricconsistentexplorable} similarly uses generated panoramic content to construct a geometrically consistent Gaussian scaffold that guides subsequent view generation. Other approaches explore related directions, including camera-controlled video generation in DimensionX~\cite{Sun_2025_ICCV}, instance-aware depth-guided reconstruction in DepR~\cite{zhao2025deprdepthguidedsingleview}, and direct diffusion-based generation of Gaussian representations in DiffusionGS~\cite{cai2025bakinggaussiansplattingdiffusion}.

VistaDream~\cite{vistadream} is particularly related to our setting. It starts from a zoomed-out and inpainted image, estimates depth, and builds an initial 3D scaffold before progressively completing the scene with additional RGB-D observations. The resulting Gaussian representation is further refined using multi-view consistency sampling. RealmDreamer~\cite{shriram2025realmdreamertextdriven3dscene} also combines Gaussian scene representations with image inpainting and depth-based guidance, although its primary focus is text-driven 3D scene generation.

\subsection{Generative Scene Completion}

Novel viewpoints often reveal regions that were not visible in the original observation. Generating plausible content for such regions is only part of the problem: the newly synthesized appearance must also remain consistent with the existing geometry and with neighboring viewpoints.

Existing methods address this issue in different ways. VistaDream~\cite{vistadream} repeatedly extends an RGB-D reconstruction and uses multi-view consistency sampling to refine the generated observations. ExScene~\cite{gong2025exscenefreeview3dscene} combines generative priors with an explicit 3D Gaussian representation, while One2Scene~\cite{wang2026one2scenegeometricconsistentexplorable} conditions further view generation on an already reconstructed Gaussian scaffold. Related approaches such as 3D-Adapter~\cite{chen20253dadaptergeometryconsistentmultiviewdiffusion} introduce explicit geometric feedback directly into the generation process.

In contrast, ORCA uses generation only when the current reconstruction does not provide sufficient information to recover the missing region. Small disocclusions are repaired directly from existing RGB-D scene information, while generative completion is reserved for larger unseen regions.

\section{Method}
\label{sec:method}
\begin{figure*}[t]
    \centering
    \includegraphics[width=0.85\linewidth]{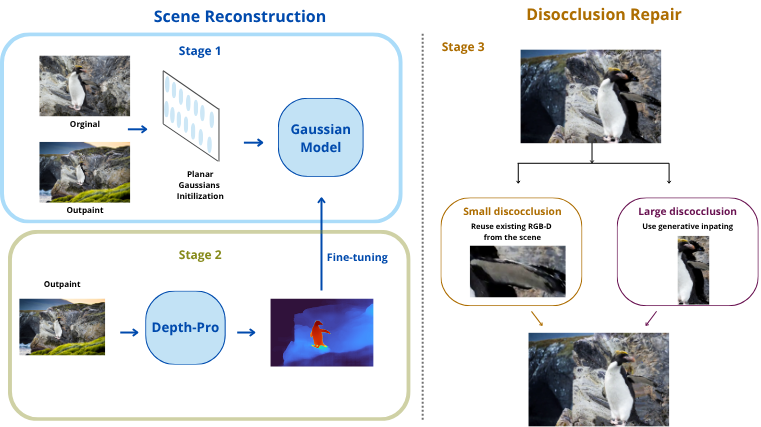}
    \caption{
        Starting from the original and outpainted images, we initialize and optimize a planar Gaussian scene representation. Monocular depth is then used to deform the learned Gaussian geometry, and the resulting representation is fine-tuned while keeping the deformed Gaussian positions fixed. During novel-view exploration, small disocclusions are repaired using RGB-D information already available in the scene, whereas larger unseen regions are completed with generative inpainting.
    }
    \label{fig:method_overview}
\end{figure*}
Our pipeline consists of three stages, as shown in Figure \ref{fig:method_overview}. We first optimize an initial Gaussian representation using the input image and its outpainted extension. We then deform the Gaussian geometry according to estimated monocular depth and fine-tune the resulting representation. Finally, we reconstruct regions revealed by viewpoint changes, with particular focus on previously occluded areas.

\subsection{Initial Scene Representation}

Given an input image \(I_{\mathrm{original}}\), we generate an outpainted version \(I_{\mathrm{outpaint}}\). The planar Gaussian cloud is initialized from the outpainted image grid, while both the original and outpainted images are used as training observations. No depth information is used at this stage. The Gaussian means are initially placed on a planar grid corresponding to the image plane.

We represent the scene using neural anchors parameterized by 3D Gaussians, following the base architecture~\cite{wilczynski2026irisintersectionawareraybasedimplicit}. The representation is defined as:

\begin{equation}
\boldsymbol{\mathcal{P}}
=
\left\{
\left(
\boldsymbol{\mathcal{N}}
(\boldsymbol{\mu}_{i},\boldsymbol{\Sigma}_{i}),
\omega_{i},
\boldsymbol{f}_{i}
\right)
\right\}_{i=1}^{N}.
\end{equation}

Here, \(\boldsymbol{\mu}_i\) and \(\boldsymbol{\Sigma}_i\) denote the mean and covariance of the \(i\)-th Gaussian, \(\omega_i\) is a logit controlling its opacity, and \(\boldsymbol{f}_i\) is its latent feature. During training, latent features are obtained from a learnable multi-resolution hash encoding queried at the Gaussian positions. During rendering, the Ray Intersection Selector identifies ray--Gaussian interactions. Features associated with neighboring intersected anchors are aggregated by the Ray-Coherent Aggregation module and decoded into density and view-dependent color, which are then integrated using volumetric rendering.

Both \(I_{\mathrm{original}}\) and \(I_{\mathrm{outpaint}}\) are assigned the same camera pose. For the outpainted image, the camera intrinsics are adjusted to account for the scale and position of the original image within the extended canvas. Because both observations share the same camera center and orientation, they provide no multi-view parallax and therefore do not constrain scene depth.

The first stage yields a Gaussian representation consistent with both training images, but its depth remains underconstrained because both observations share the same viewpoint. We address this in the second stage using monocular depth.

\subsection{Depth-Based Gaussian Deformation}

To introduce depth structure, we estimate a depth map \(D\) from the outpainted image \(I_{\mathrm{outpaint}}\) using Depth Pro~\cite{bochkovskii2025depthprosharpmonocular}. For each Gaussian mean, its \(x\)- and \(y\)-coordinates are independently normalized with respect to the spatial extent of the current Gaussian cloud and mapped to normalized image coordinates. The depth map is then bilinearly sampled at the corresponding location.

Rather than using the metric depth values directly, we convert them to inverse depth and normalize them into a relative nearness measure. This allows us to use the predicted relative scene structure while adapting the magnitude of the depth deformation to the scale of the Gaussian cloud. Let

\begin{equation}
q_i = \frac{1}{\max(d_i,\epsilon)},
\end{equation}

where \(d_i\) is the depth value bilinearly sampled for the \(i\)-th Gaussian and \(\epsilon\) is a small positive constant preventing division by zero.
Larger \(q_i\) corresponds to points closer to the camera. We then compute

\begin{equation}
s_i =
\operatorname{clip}
\left(
\frac{q_i-q_{\mathrm{far}}}
{q_{\mathrm{near}}-q_{\mathrm{far}}},
0,1
\right),
\end{equation}

where \(q_{\mathrm{far}}\) and \(q_{\mathrm{near}}\) are robust lower and upper quantiles of the inverse-depth distribution. This step reduces the influence of extreme depth predictions and expresses the estimated geometry as a relative displacement within the scale of the Gaussian representation.

The normalized value \(s_i\) determines the target \(z\)-coordinate of each Gaussian:

\begin{equation}
\tilde{z}_i =
z_{\mathrm{base}}
+
\lambda L_{xy}s_i,
\end{equation}

where \(z_{\mathrm{base}}\) is the median \(z\)-coordinate of the current Gaussian cloud,

\begin{equation}
L_{xy}
=
\max
\left(
x_{\max}-x_{\min},
y_{\max}-y_{\min}
\right),
\end{equation}

is the maximum spatial extent of the cloud along the \(x\)- and \(y\)-axes, and \(\lambda\) controls the magnitude of the deformation.

Changing only the \(z\)-coordinate would move a Gaussian away from its original viewing ray. We therefore displace each Gaussian along the ray passing through its current position. Let \(\mathbf{c}\) denote the camera center, with \(c_z\) its \(z\)-coordinate, and let \(\mu_{i,z}\) denote the current \(z\)-coordinate of the \(i\)-th Gaussian mean. We compute:

\begin{equation}
\gamma_i =
\frac{c_z-\tilde{z}_i}
{c_z-\mu_{i,z}},
\end{equation}

and update the Gaussian mean as:

\begin{equation}
\tilde{\boldsymbol{\mu}}_i
=
\mathbf{c}
+
\gamma_i
\left(
\boldsymbol{\mu}_i-\mathbf{c}
\right).
\end{equation}

This construction places the Gaussian at the target \(z\)-coordinate while preserving its direction from the camera center. The image-space correspondence established during the first stage is therefore retained while the Gaussian cloud is deformed according to the estimated scene geometry.

The Stage-1 checkpoint is then fine-tuned using the deformed Gaussian positions. We replace the original means with \(\tilde{\boldsymbol{\mu}}_i\) and keep them fixed during this stage, while the remaining model parameters continue to be optimized. Fine-tuning uses the same RGB reconstruction objective as in the first stage, without introducing an additional depth loss. The depth estimate is used only to modify the Gaussian geometry before fine-tuning and is not used as a direct supervision signal.

\subsection{Disocclusion Repair}

After depth-guided fine-tuning, moving the camera can reveal regions that are not covered by the existing Gaussian cloud. These regions usually appear as thin gaps near foreground--background boundaries or as larger areas that were not visible in the input image. The third stage detects and repairs these disocclusions.

We render the reconstructed scene along a predefined camera trajectory. At each iteration, we select the view with the largest detected disocclusion and split its repair mask into connected components. Small or elongated components are treated as local geometric gaps, while large and compact components are considered for generative completion. Local components are processed first, since many of these gaps can be repaired using information already present in the reconstructed scene. 

For each local component, we build an RGB-D target using the surrounding reconstructed background. We first extract a ring of valid pixels around the repair region and select donors from the farther part of the local depth distribution. Using the nearest depth at an occlusion boundary can incorrectly extend the foreground surface into the missing region. Selecting farther background samples reduces this effect and places the added geometry behind the foreground object. RGB and depth values for the repair region are taken from nearby valid donors. If not enough valid background samples are available, we use nearest-neighbor filling instead. 

New Gaussian anchors are initialized from the resulting RGB-D target and added only in the missing region. The newly added Gaussians are then optimized locally, while the existing scene representation remains unchanged. The optimization matches the target RGB values, increases coverage inside the repaired region, and preserves the appearance around its boundary. We use wider, overlapping Gaussians for these local repairs to reduce thin gaps that can remain visible after a viewpoint change. 

Some regions cannot be recovered reliably from the local background. If a remaining component is sufficiently large and compact, we inpaint it using Stable Diffusion. Inpainting is applied to a crop around the selected component rather than to the whole rendered image, and only pixels inside the missing region are replaced. We allow at most two generative inpainting operations per reconstructed scene. After this budget is exhausted, remaining regions are handled using the local geometry-based repair. 

For generated regions, we estimate depth using Depth Pro. The predicted depth is aligned with the depth of the current reconstruction using a local affine transformation,

\begin{equation}
d_{\mathrm{aligned}}
=
a d_{\mathrm{pred}} + b,
\end{equation}

where \(a\) and \(b\) are estimated from valid background pixels around the repaired region. Outliers are removed during the fitting. Near the boundary of the repaired region, the aligned depth is blended with the reconstructed background depth. 
The completed RGB-D region is then converted into additional Gaussian anchors and optimized locally. If inpainting or depth estimation fails, we fall back to the local background-based repair. 
After each repair, we render the camera trajectory again and search for the next view with remaining disocclusions. The newly added Gaussians are therefore evaluated from the full trajectory before another region is selected. The procedure stops when no sufficiently large disocclusion remains or when the maximum number of repair iterations is reached.

\section{Experiments}
\label{sec:experiments}

We evaluate ORCA from three complementary perspectives: the perceptual quality of rendered novel views, cross-view geometric consistency, and qualitative behavior under camera motion. We compare against VistaDream on two datasets using matched camera trajectories for both methods.

\subsection{Datasets}
We evaluate our method on two sets: DIV2K~\cite{Timofte_2017_CVPR_Workshops} and a subset of scenes released with RealmDreamer~\cite{shriram2025realmdreamertextdriven3dscene}. DIV2K provides a larger and more diverse collection of natural images, while the RealmDreamer subset provides a smaller set of synthetic scenes.
For DIV2K, we use the validation split of the \(2\times\) bicubic downsampling setting. It contains 100 images obtained by bicubic downsampling of the corresponding high-resolution images, with spatial resolutions ranging from \(408\times1020\) to \(1020\times1020\). The variety of scenes and image content allows us to evaluate the method on a broad set of natural inputs. VistaDream failed to produce a valid reconstruction for one sky-dominated image in the DIV2K validation set. We therefore exclude this scene from the comparison for both methods and report quantitative results on the remaining 99 images.
We additionally evaluate on 11 scenes from RealmDreamer: \textit{bathroom}, \textit{bear}, \textit{bedroom3}, \textit{bust}, \textit{car}, \textit{kitchen}, \textit{lavender}, \textit{living\_room}, \textit{piano}, \textit{steampunk}, and \textit{victorian}. All images have a resolution of \(512\times512\) pixels. These scenes include both indoor environments and object-centered compositions.

\subsection{Implementation Details}

The initial Gaussian representation is trained for at most \(30{,}000\) iterations. Early stopping is based on a smoothed PSNR score evaluated every \(1{,}000\) iterations, with a minimum improvement of \(0.03\) dB and a patience of four evaluations.

After the initial training stage, Depth Pro is applied to the outpainted image and the resulting depth estimate is used to deform the Gaussian positions as described in Sec.~\ref{sec:method}. The depth-guided representation is then fine-tuned for at most \(20{,}000\) iterations. Early stopping is enabled after \(10{,}000\) iterations, using the same stopping parameters as in the first stage. During fine-tuning, the deformed Gaussian means remain fixed while the remaining model parameters are optimized.

For disocclusion repair, local geometry-based completion is used by default. A missing region is considered for generative completion only if it contains at least \(9{,}000\) pixels, occupies at least \(45\%\) of its bounding box, and the shorter side of the bounding box is at least \(64\) pixels. All other components are repaired using information from the surrounding reconstructed background.

For local repair, RGB-D donor pixels are searched within a ring around the missing region, with inner and outer radii of \(3\) and \(28\) pixels, respectively. We require at least \(32\) valid donor pixels and preferentially select samples from the farther part of the local depth distribution, using the \(78\)-th depth percentile as the default threshold. Newly inserted Gaussians use a covariance scale multiplier of \(1.7\) and a minimum opacity of \(0.9\). At most \(5{,}000\) Gaussians are added in a single local repair, and only the newly inserted Gaussians are optimized, for up to \(24\) steps.

For generative completion, we use the Stable Diffusion 1.5 inpainting model \cite{Rombach_2022_CVPR}. Inpainting is performed on a crop around the selected missing region using \(30\) inference steps, a guidance scale of \(7.0\), and a maximum resolution of \(512\) pixels. At most two generative completion operations are allowed per reconstructed scene. For generated regions, Depth Pro is used to estimate depth, which is locally aligned to the depth of the current reconstruction before the corresponding Gaussian anchors are inserted.

All reported experiments were run on a single NVIDIA A40 GPU. We additionally verified that the full pipeline can be executed on an NVIDIA RTX 4060 GPU.

\subsection{Quantitative Results}
We compare our method with VistaDream~\cite{vistadream} on both DIV2K and RealmDreamer datasets. For both methods, we evaluate rendered camera trajectories using the same set of image-quality metrics. To ensure a fair comparison, we render VistaDream ~\cite{vistadream} using the same camera trajectories as those used for our method. We report MUSIQ \cite{ke2021musiqmultiscaleimagequality} and CLIP-IQA \cite{wang2022exploringclipassessinglook}, together with the five LLaVA-IQA criteria used in VistaDream~\cite{vistadream} : Noise-Free (NF), Edge, Structure, Detail, and overall Quality. For each reconstructed scene, the metrics are averaged over 50 frames sampled from the rendered video. We then report the mean score over all scenes in each dataset. Table~\ref{tab:quantitative_comparison} summarizes the results on DIV2K and RealmDreamer. ORCA consistently outperforms VistaDream across all reported metrics on both datasets. On DIV2K, MUSIQ increases from \(61.60\) to \(68.71\), while CLIP-IQA improves from \(0.474\) to \(0.574\). The LLaVA-IQA scores show a similar trend, with overall Quality increasing from \(0.407\) to \(0.630\), together with substantial improvements in Edge and Structure. On RealmDreamer, ORCA improves MUSIQ from \(68.66\) to \(72.85\) and CLIP-IQA from \(0.378\) to \(0.457\), while overall Quality increases from \(0.573\) to \(0.851\). These results indicate that the proposed reconstruction and repair strategy improves both perceptual image quality and the preservation of scene structure during novel-view exploration.

\begin{table*}[h]
    \centering
    \caption{Quantitative comparison with VistaDream on DIV2K and RealmDreamer. Results are averaged over all evaluated scenes in each dataset. Higher values are better for all metrics.}
    \label{tab:quantitative_comparison}
    \resizebox{\textwidth}{!}{
    \begin{tabular}{llccccccc}
        \hline
        Dataset & Method &
        MUSIQ $\uparrow$ &
        CLIP-IQA $\uparrow$ &
        NF $\uparrow$ &
        Edge $\uparrow$ &
        Structure $\uparrow$ &
        Detail $\uparrow$ &
        Quality $\uparrow$ \\
        \hline

        \multirow{2}{*}{DIV2K}
        & VistaDream & 61.6023 & 0.4735 & 0.5808 & 0.3663 & 0.4337 & 0.7206 & 0.4065 \\
        & Ours       & \textbf{68.7133} & \textbf{0.5735} & \textbf{0.7606} & \textbf{0.5980} & \textbf{0.6539} & \textbf{0.8962} & \textbf{0.6295} \\
        \hline

    \multirow{2}{*}{RealmDreamer}
    & VistaDream & 68.6576 & 0.3775 & 0.7982 & 0.4291 & 0.4818 & 0.9327 & 0.5727 \\
    & Ours       & \textbf{72.8514} & \textbf{0.4567} & \textbf{0.9636} & \textbf{0.7182} & \textbf{0.6855} & \textbf{1.0000} & \textbf{0.8509} \\
    \hline
    \end{tabular}
    }
\end{table*}

%noise_free — „Czy obraz jest wolny od szumu lub zniekształceń?”
%edge — „Czy obraz pokazuje wyraźne obiekty i ostre krawędzie?”
%structure — „Czy ogólna scena jest spójna i realistyczna pod względem układu i proporcji?”
%detail — „Czy obraz pokazuje szczegółowe tekstury i materiały?”
%quality — „Czy obraz jest ogólnie wysokiej jakości, z wyraźnymi obiektami, ostrymi krawędziami, dobrymi kolorami, dobrą strukturą i dobrą jakością wizualną?”

Since RealmDreamer contains a small set of predefined scenes, we additionally report scene-level results in Table~\ref{tab:realmdreamer_per_scene}. This allows us to examine how the two methods perform across different scene types rather than relying only on the average score. The per-scene results show that the improvement is consistent across different scene types. ORCA achieves higher MUSIQ and CLIP-IQA scores than VistaDream on all 11 RealmDreamer scenes. The largest improvements are observed for several of the more challenging scenes, including \textit{bear} and \textit{piano}, where VistaDream exhibits substantial degradation under viewpoint changes. While individual LLaVA-IQA criteria occasionally remain unchanged or favor VistaDream, the overall trend strongly favors ORCA. 

\begin{table*}[h]
    \centering
    \caption{Per-scene comparison with VistaDream on the 11 RealmDreamer scenes. Higher values are better for all metrics.}
    \label{tab:realmdreamer_per_scene}
    \resizebox{\textwidth}{!}{
    \begin{tabular}{llccccccc}
        \hline
        Scene & Method &
        MUSIQ $\uparrow$ &
        CLIP-IQA $\uparrow$ &
        NF $\uparrow$ &
        Edge $\uparrow$ &
        Structure $\uparrow$ &
        Detail $\uparrow$ &
        Quality $\uparrow$ \\
        \hline

        \multirow{2}{*}{bathroom}
        & VistaDream & 70.8680 & 0.4333 & 1.00 & 0.60 & 0.70 & 1.00 & 0.92 \\
        & Ours       & \textbf{73.8245} & \textbf{0.4643} & 1.00 & \textbf{1.00} & \textbf{0.98} & 1.00 & \textbf{1.00} \\

        \multirow{2}{*}{bear}
        & VistaDream & 61.5041 & 0.4851 & 0.38 & 0.22 & 0.00 & 0.52 & 0.22 \\
        & Ours       & \textbf{71.0969} & \textbf{0.6452} & \textbf{0.96} & \textbf{0.80} & 0.00 & \textbf{1.00} & \textbf{0.68} \\

        \multirow{2}{*}{bedroom3}
        & VistaDream & 72.8725 & 0.2940 & 0.96 & 0.70 & 0.90 & 1.00 & 0.88 \\
        & Ours       & \textbf{75.0659} & \textbf{0.3245} & \textbf{1.00} & \textbf{0.96} & \textbf{1.00} & 1.00 & \textbf{1.00} \\

        \multirow{2}{*}{bust}
        & VistaDream & 71.8854 & 0.3467 & 0.90 & 0.46 & 0.54 & 1.00 & 0.54 \\
        & Ours       & \textbf{75.7703} & \textbf{0.4273} & \textbf{1.00} & \textbf{1.00} & \textbf{1.00} & 1.00 & \textbf{1.00} \\

        \multirow{2}{*}{car}
        & VistaDream & 58.1490 & 0.4131 & 0.30 & 0.00 & 0.00 & 0.88 & \textbf{0.02} \\
        & Ours       & \textbf{64.8457} & \textbf{0.4805} & \textbf{0.64} & 0.00 & 0.00 & \textbf{1.00} & 0.00 \\

        \multirow{2}{*}{kitchen}
        & VistaDream & 64.4678 & 0.3558 & 1.00 & 0.80 & 1.00 & 1.00 & 0.74 \\
        & Ours       & \textbf{69.6226} & \textbf{0.4644} & 1.00 & \textbf{1.00} & 1.00 & 1.00 & \textbf{0.96} \\

        \multirow{2}{*}{lavender}
        & VistaDream & 66.6912 & 0.3659 & 1.00 & 0.72 & 0.52 & 1.00 & 0.42 \\
        & Ours       & \textbf{70.2838} & \textbf{0.3958} & 1.00 & \textbf{0.78} & \textbf{0.70} & 1.00 & \textbf{0.78} \\

        \multirow{2}{*}{living\_room}
        & VistaDream & 71.8811 & 0.2763 & 0.96 & 0.58 & 0.80 & 0.90 & 0.74 \\
        & Ours       & \textbf{75.7188} & \textbf{0.3015} & \textbf{1.00} & \textbf{1.00} & \textbf{1.00} & \textbf{1.00} & \textbf{1.00} \\

        \multirow{2}{*}{piano}
        & VistaDream & 69.4828 & 0.3858 & 0.56 & 0.38 & 0.38 & 0.96 & 0.38 \\
        & Ours       & \textbf{75.1793} & \textbf{0.5491} & \textbf{1.00} & \textbf{0.98} & \textbf{1.00} & \textbf{1.00} & \textbf{0.94} \\

        \multirow{2}{*}{steampunk}
        & VistaDream & 75.8560 & 0.4399 & 1.00 & 0.00 & 0.00 & 1.00 & 0.96 \\
        & Ours       & \textbf{75.9137} & \textbf{0.5457} & 1.00 & 0.00 & 0.00 & 1.00 & \textbf{1.00} \\

        \multirow{2}{*}{victorian}
        & VistaDream & 71.5758 & 0.3568 & 0.72 & 0.26 & 0.46 & 1.00 & 0.48 \\
        & Ours       & \textbf{74.0436} & \textbf{0.4259} & \textbf{1.00} & \textbf{0.38} & \textbf{0.86} & 1.00 & \textbf{1.00} \\

        \hline
    \end{tabular}
    }
\end{table*}

We additionally evaluate cross-view geometric consistency using TSED~\cite{yu2023longtermphotometricconsistentnovel}. Unlike the image-quality metrics reported above, TSED measures whether correspondences across rendered views are consistent with the underlying camera geometry. The results are reported in Table~\ref{tab:tsed_comparison}. ORCA also achieves higher cross-view geometric consistency. On DIV2K, TSED increases from \(0.8265\) for VistaDream to \(0.9980\) for ORCA, while on RealmDreamer it increases from \(0.9864\) to \(1.0000\). The particularly large improvement on DIV2K indicates that the gains in perceptual quality are accompanied by substantially higher cross-view consistency rather than being limited to individual rendered frames. The improvements across both perceptual and geometric metrics suggest that ORCA improves not only individual frame quality but also the stability of the reconstructed scene across viewpoints.

\subsection{Qualitative Results}
\begin{figure*}[t]
    \centering
    \includegraphics[width=\textwidth]{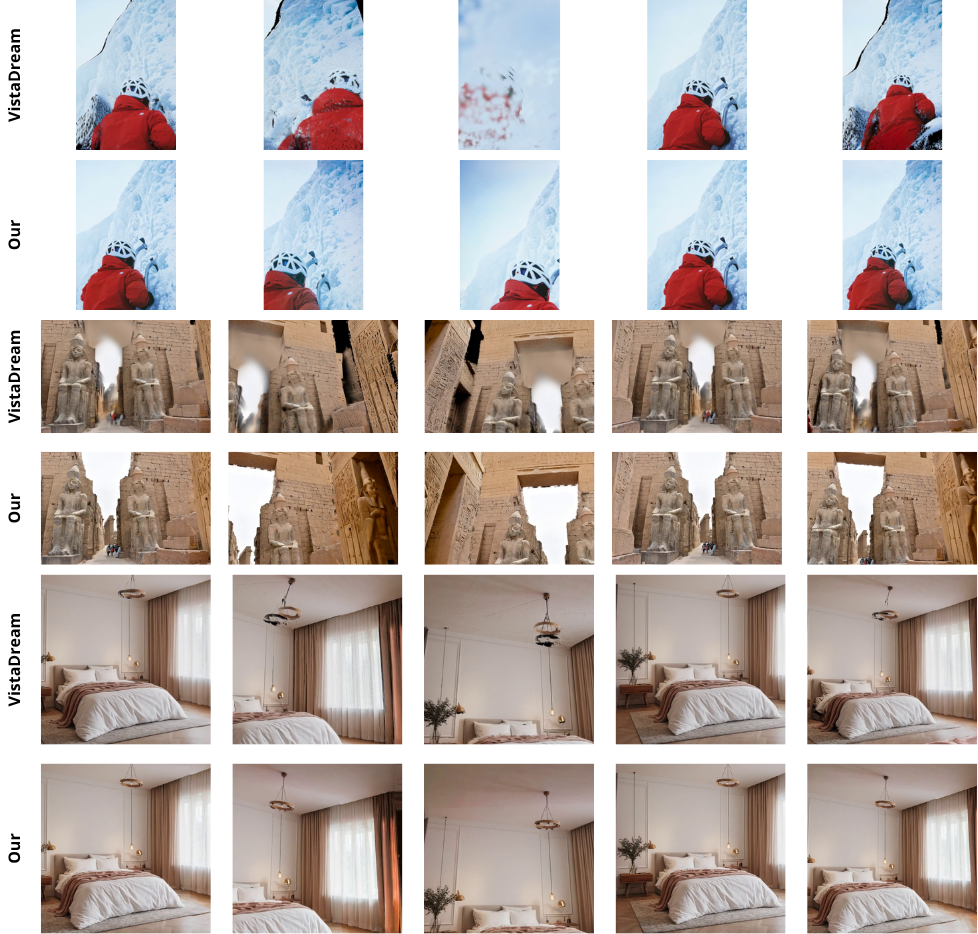}
    \caption{\textbf{Qualitative comparison with VistaDream.}
    Each pair of rows shows matched novel views rendered by VistaDream (top)
    and ORCA (bottom) along the same camera trajectory. VistaDream exhibits
    uncovered regions, Gaussian-like artifacts, and blurring in challenging
    viewpoints, whereas ORCA produces more complete and spatially coherent
    renderings while better preserving scene structure across views.}
    \label{fig:qualitative_comparison}
\end{figure*}

Figure~\ref{fig:qualitative_comparison} presents representative novel views
rendered along matched camera trajectories. The qualitative results are
consistent with the quantitative evaluation. As the camera moves away from the
input viewpoint, VistaDream can exhibit uncovered regions, local Gaussian-like
artifacts, and substantial blurring in newly exposed areas. These artifacts are
particularly visible near occlusion boundaries and in views requiring larger
viewpoint changes.

ORCA produces more complete and visually stable novel views. Small
disocclusions are filled without unnecessarily changing the surrounding
appearance, while larger missing regions are integrated into the reconstructed
scene without introducing the same degree of visible degradation. Across the
shown trajectories, scene structure and object boundaries remain more consistent
as the viewpoint changes.

\begin{table}[ht]
    \centering
    \caption{Geometric consistency measured with TSED on DIV2K and RealmDreamer. Higher values indicate better cross-view geometric consistency.}
    \label{tab:tsed_comparison}
    \begin{tabular}{lcc}
        \hline
        Dataset & VistaDream & Ours \\
        \hline
        DIV2K        & 0.8265 & \textbf{0.9980} \\
        RealmDreamer & 0.9864 & \textbf{1.0000} \\
        \hline
    \end{tabular}
\end{table}

\section{Conclusion}
\label{sec:conclusion}

We presented ORCA, an occlusion-aware method for reconstructing explorable 3D scenes from a single image. Starting from a planar Gaussian representation, ORCA introduces scene depth using monocular depth estimates while preserving the original camera-ray correspondence. During novel-view exploration, the method treats small disocclusions differently from regions containing genuinely unseen content. Small gaps are repaired using RGB-D information already present in the reconstruction, while generative inpainting is used only when the missing region cannot be recovered reliably from the scene. Newly added geometry is optimized locally without modifying the existing representation.

Experiments on DIV2K and RealmDreamer show that ORCA improves both perceptual novel-view quality and cross-view geometric consistency compared with VistaDream. The results support the main motivation of our approach: many artifacts revealed by camera motion can be repaired using information already available in the reconstructed scene, reducing the need to generate new content.

{
    \small
    \bibliographystyle{ieeenat_fullname}
    \bibliography{main}
}

\end{document}